\documentclass[a4paper, 10 pt, conference]{IEEEtran}
\IEEEoverridecommandlockouts

\usepackage{cite}

\usepackage{amsmath,amssymb,amsfonts}
\usepackage{algorithmic}
\usepackage{graphicx}
\usepackage{textcomp}
\usepackage{xcolor}
\usepackage{breqn}
\usepackage{makecell}
\usepackage{authblk}
\usepackage{inputenc}

\newcommand{\RomanNumeralCaps}[1]
    {\MakeUppercase{\romannumeral #1}}

\def\BibTeX{{\rm B\kern-.05em{\sc i\kern-.025em b}\kern-.08em
    T\kern-.1667em\lower.7ex\hbox{E}\kern-.125emX}}
\begin{document}

\title{Affine Disentangled GAN for Interpretable and Robust AV Perception\\
{\footnotesize \textsuperscript{}}
\thanks{This work is funded by Industrial Postgraduate Program (IPP). Serial Number: S17-1298-IPP-II, Singapore.}
}

\author[12]{Liu Letao} 
\author[2]{Martin Saerbeck}
\author[1]{Justin Dauwels}
\affil[1]{School of Electrical and Electronic Engineering \protect\\Nanyang Technological University, 50 Nanyang Avenue, Singapore, 639798}
\affil[2]{TUV SUD Singapore \protect\\1 Science Park Dr, Singapore, 118221}

\maketitle

\begin{abstract}
Autonomous vehicles (AV) have progressed rapidly with the advancements in computer vision algorithms. The deep convolutional neural network as the main contributor to this advancement has boosted the classification accuracy dramatically. However, the discovery of adversarial examples reveals the generalization gap between dataset and the real world. Furthermore, affine transformations may also confuse computer vision based object detectors. The degradation of the perception system is undesirable for safety critical systems such as autonomous vehicles.  In this paper, a deep learning system is proposed: Affine Disentangled GAN (ADIS-GAN),  which is robust against affine transformations and adversarial attacks. It is demonstrated that conventional data augmentation for affine transformation and adversarial attacks are orthogonal, while ADIS-GAN can handle both attacks at the same time. Useful information such as image rotation angle and scaling factor are also generated in ADIS-GAN. On MNIST dataset, ADIS-GAN can achieve over 98 percent classification accuracy within 30 degrees rotation, and over 90 percent classification accuracy against FGSM and PGD adversarial attack.
\end{abstract}


\section{Introduction \& Related Work}
{Autonomous vehicles (AV) have received much attention in recent years. One pillar of AV perception systems is the RGB data captured by the camera. Through the RGB data, the system can understand its surrounding environment, including the location of vehicles, pedestrians and other crucial information. The deep convolutional neural network (CNN) is a widely accepted cutting-edge computer vision algorithm \cite{Krizhevsky:2012:ICD:2999134.2999257} \cite{DBLP:journals/corr/HeZRS15} \cite{DBLP:journals/corr/SzegedyVISW15} to process the RGB data, for detecting objects, segmenting urban scenes, etc. 
Despite the tremendous success accomplished by deep CNN models, the adversarial examples show that there are still reliability and robustness issues (see Fig.~1). The adversarial images are visually indistinguishable for a human viewer, but state-of-the-art classifiers make wrong predictions with high confidence for those images. Since the first publication of adversarial attack \cite{DBLP:journals/corr/SzegedyZSBEGF13}, numerous studies have appeared on this topic \cite{DBLP:journals/corr/GoodfellowSS14, DBLP:journals/corr/KurakinGB16, DBLP:conf/ccs/SharifBBR16, DBLP:journals/corr/TramerKPBM17, DBLP:journals/corr/Moosavi-Dezfooli15, DBLP:journals/corr/CarliniW16a, DBLP:journals/corr/PapernotMGJCS16, DBLP:journals/corr/MadryMSTV17, DBLP:journals/corr/AthalyeS17}, which mainly consider pixel level perturbations. Such adversarial attacks on CNN raise concerns that the perception system of the autonomous vehicle can be maliciously hacked. \par}

\begin{figure}[htbp]
\centerline{\includegraphics[width = 0.48\textwidth]{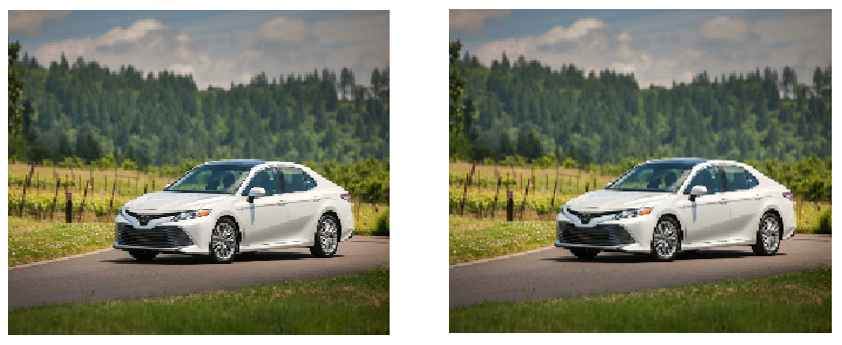}}
\caption{An adversarial attack on a car image. Left: without adversarial attack, the classification result is ``minivan''. Right: with FGSM adversarial attack, the classifier labels this image as ``car wheel'' instead of ``minivan''. Testing algorithm: ResNet-50 \cite{DBLP:journals/corr/HeZRS15}.}
\end{figure}

{In \cite{DBLP:journals/corr/abs-1712-02779}, it was shown that simple affine transformations (e.g., rotations) can cause deep CNNs to misclassify images.  The images captured by the perception system could experience similar affine distortions during normal driving scenarios when vehicles are passing water puddles (see Fig.~2) or on rural roads. Both adversarial attack and affine distortions need to be addressed before integrating deep CNN vision systems into safety-related applications like autonomous vehicle at large scale. \par }

\begin{figure}[htbp]
\centerline{\includegraphics[width = 0.35\textwidth]{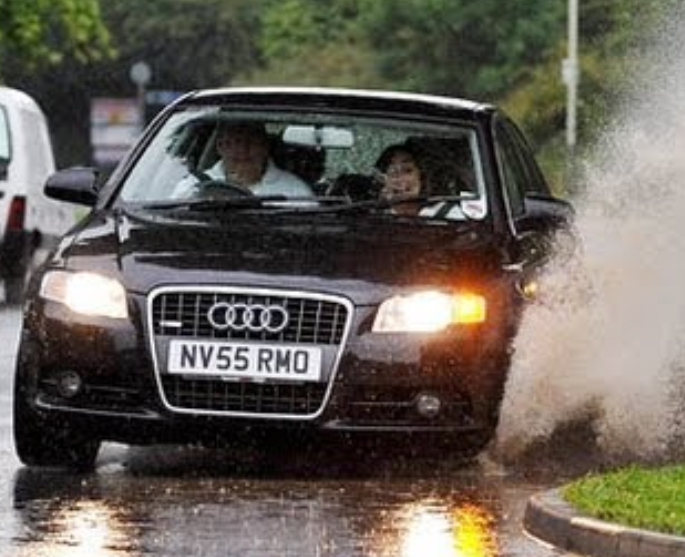}}
\caption{When the vehicle hits a water puddle, the images captured by the camera will be tilted. As a result, the RGB-based object detector may misclassify objects in scenes.}
\end{figure}

\begin{figure}[htbp]
\centerline{\includegraphics[width = 0.45\textwidth]{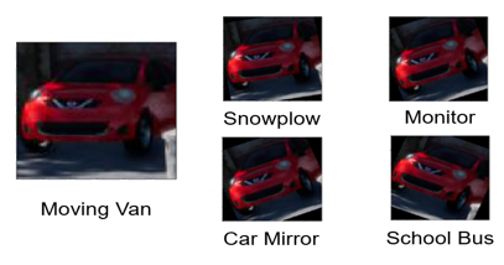}}
\caption{Rotation of vehicle images. The vehicle tends to be detected incorrectly when the image is tilted. Testing algorithm: Inception v3 \cite{DBLP:journals/corr/SzegedyVISW15}.}
\end{figure}

{GAN \cite{NIPS2014_5423} has been widely studied and utilized since its invention. It is a generative model which captures high-dimensional data distribution through adversarial process. It can generate images that simulate the training images such as hand written digits, animals, vehicles, etc. Deep Convolutional GAN \cite{DBLP:journals/corr/RadfordMC15} introduces convolution mechanism into the GAN structure by inserting the deconvolution layers in the generator network. Bi-directional GAN \cite{DBLP:journals/corr/DonahueKD16} further provides a pathway to convert data from image space back to latent space with additional encoder network. InfoGAN \cite{DBLP:journals/corr/ChenDHSSA16} utilizes a disentangled representation that separates features and noise in the latent space. The separated features can represent categorical and continuous attributes of the training images. In \cite{DBLP:journals/corr/abs-1811-12359}, the issue of inductive bias in disentangled representation is discussed. In \cite{DBLP:journals/corr/abs-1812-02230}, the concept of symmetry group concept is introduced to define disentanglement behaviour. DefenceGAN \cite{DBLP:journals/corr/abs-1805-06605} uses GAN as a defence method against adversarial attacks.}

{Several studies address invariance property of deep CNN for affine transformations. In \cite{DBLP:journals/corr/JaderbergSZK15}, anti-distortion classification result is achieved by inserting Spatial Transformer layer into the given network. However the affine parameters are not presented in a disentangled manner, which makes it less interpretable. Transforming Autoencoder\cite{Hinton:2011:TA:2029556.2029562} uses auto-encoder to model 2D affine transformation applied to images. The trained generative model can learn to generate transformed images in a disentangled way. However it does not tackle the improvement on classification accuracy. In \cite{NIPS2014_5424}, transformations that preserve the object identity are analyzed in the symmetry group. In \cite{DBLP:journals/corr/KanazawaSJ14}\cite{Sohn:2012:LIR:3042573.3042745}, filter banks are designed to make the classifier transformation invariant. }

\hfill \break
{\noindent}\textbf{Our Contributions}
\begin{enumerate}
\item{We introduce Affine Disentangled GAN (ADIS-GAN) which is robust against both affine transformation and adversarial attack. It achieves classification accuracies comparable to that of state-of-the-art supervised learning algorithm, although it is an unsupervised algorithm. }
\item{We show that affine transformation augmented training and adversarial augmented training are orthogonal, which means they can only defend typical attack they have been trained on.}
\item{Affine Disentangled GAN is more interpretable, providing information that helps to understand potential misclassifications. On MNIST dataset, it can achieve over 98 percent classification accuracy within 30 degrees rotation, and over 90 percent classification accuracy against FGSM and PGD adversarial attack. }
\end{enumerate}

\section{Preliminary}
\subsection{Generative Adversarial Network}
{GAN \cite{NIPS2014_5423} is a generative model which captures high-dimensional data distribution through adversarial process: a mini-max game between the generator and discriminator. The generator tries to produce images that are similar to real ones, while the discriminator judges whether the images are generated or real. During the training process, the generator will create images that do not belong to the original dataset. Those images may prevent the model from overfitting, and the model is more likely to learn a smoother data distribution which involves the adversarial samples. The vanilla GAN formulation is: \par}

\begin{equation}
\begin{split}
\min_{G}\max_{D}V(D,G) := & \mathbb{E}_{x\sim{P_{\text{data}}(x)}}\log[D(x)] \\
&+ \mathbb{E}_{z\sim{P_{z}(z)}}\log[1-D(G(z))].
\end{split}
\end{equation}

{A standard distribution $z\sim P_z$ in the latent space can be transferred to data space $P_x$ through generator $G$. The discriminator $D$ judges whether the samples are from training dataset $x$ or generated dataset $G(z)$. \par}

{Bi-directional GAN \cite{DBLP:journals/corr/DonahueKD16} adds an encoder to the vanilla GAN, which makes the image to latent and latent to image transformation possible. The encoder and generator together can be treated as a filter where the reconstructed images may only keep meaningful information and discard noise such as adversarial perturbation. The Bi-directional GAN formulation is: \par}

\begin{equation}	
\begin{split}
\min_{G,E}\max_{D}V(D,G,E) := & \mathbb{E}_{x\sim{P_{\text{data}}(x)}}\log[D(x, E(x))] \\
& + \mathbb{E}_{z\sim{P_{z}(z)}}\log[1-D(z,G(z))].
\end{split}
\end{equation}

{InfoGAN \cite{DBLP:journals/corr/ChenDHSSA16} can assign the latent vectors semantic meanings such as categorical and continuous information (e.g. skew of an image) by maximizing the mutual information between generated latent space and reconstructed latent space. The InfoGAN formulation is: }

\begin{equation}
\begin{split}
&\min_{G,Q}\max_{D}V(D,G,Q) :=  \mathbb{E}_{x\sim{P_{\text{data}}(x)}}\log[D(x)] \\
& + \mathbb{E}_{{z\sim{P_{z}(z)}},{c\sim{P_{c}(c)}}}\log[1-D(G(c,z))] - \lambda I(c,Q(G(c,z))).
\end{split}
\end{equation}

{Bi-directional Info GAN \cite{DBLP:journals/corr/abs-1803-02627} uses encoder $E$ instead auxiliary $Q$ to reconstruct the latent vectors.  The Bi-directional InfoGAN formulation is : }

\begin{equation}
\begin{aligned}
&\min_{G,E}\max_{D}V(D,G,E) := \mathbb{E}_{x\sim{P_{\text{data}}(x)}}\log[D(x, E(x))] \\
& +\mathbb{E}_{{z\sim{P_{z}(z)}},{c\sim{P_{c}(c)}}}\log[1-D((c,z), G(c,z))] \\
&- \lambda I(c,E(G(c,z))).
\end{aligned}
\end{equation}

\subsection{Affine Transformation Matrix}
{Inspired by \cite{DBLP:journals/corr/JaderbergSZK15}, we utilize the affine matrix as a regularizer in our model. Conventional affine matrix is a 2 by 3 matrix, matrix defined as:}

\begin{gather}
\begin{bmatrix}A_{11} & A_{12} & A_{13} \\ A_{21} & A_{22} & A_{23}\end{bmatrix}.
\end{gather}

{$A_{13}$ and $A_{23}$ represent the horizontal and vertical translation parameters respectively. These 2 parameters can be removed from the affine matrix without affecting other affine properties. The affine matrix becomes a 2 by 2 matrix after removing the translation parameters. It can be decomposed as rotation, 
skew, and zoom matrix respectively (see Appendix for an alternative formulation):}

\begin{gather}
\begin{bmatrix}A_{11} & A_{12} \\ A_{21} & A_{22}\end{bmatrix}
=
\begin{bmatrix}\cos{\theta} & -\sin{\theta} \\ \sin{\theta} & \cos{\theta} \end{bmatrix}
\begin{bmatrix}1 & m \\ n & 1 \end{bmatrix}
\begin{bmatrix} p & 0 \\ 0 & q \end{bmatrix}.
\end{gather}

\section{System Description}

\subsection{Affine Regularizer}
{Since the images captured by the camera usually will not be skewed during normal driving scenarios, we only focus on rotation and zoom attributes in this paper. If we discard skew matrix, the affine matrix equation can be simplified as follows: }

\begin{gather}
\begin{bmatrix}A_{11} & A_{12} \\ A_{21} & A_{22}\end{bmatrix}
=
\begin{bmatrix}\cos{\theta} & -\sin{\theta} \\ \sin{\theta} & \cos{\theta} \end{bmatrix}
\begin{bmatrix} p & 0 \\ 0 & q \end{bmatrix},
\end{gather}

{where:} 

\begin{align}
& A_{11} = p(\cos{\theta}), \\
& A_{12} = q(-\sin{\theta}),  \\
& A_{21} = p(\sin{\theta}), \\
& A_{22} = q(\cos{\theta}). 
\end{align}

{Assume each image is composed of an affine matrix $M$ and a base image $I$. The input image $I_{\text{ori}}$ from training dataset can be expressed as:

\begin{align}
& I_{\text{ori}} =  M_{1} I_0. 
\end{align}

The scaled image $I_{\text{scaled}}$ transformed from $I_{\text{ori}}$ with predefined affine matrix $M_{\text{transform}}$ can be expressed as: 

\begin{align}
& I_{\text{scaled}} = M_{\text{transform}} I_{\text{ori}}.
\end{align}

With the assumption, the scaled image $I_{\text{scaled}}$ can also be expressed as:}

\begin{align}
& I_{\text{scaled}} =  M_{2} I_0. 
\end{align}

{Through simple matrix manipulation we can obtain the affine regularizer:}

\begin{align}
&M_{\text{transform}} = M_2 (M_1)^{-1}.
\end{align}

\subsection{Model Architecture}
{The Affine Disentangled GAN (ADIS-GAN) maximizes the mutual information between generated affine matrix and reconstructed affine matrix with the assumption of affine regularizer (see Fig.~4). Three continuous latent vectors are assigned to $p$, $q$, and $\theta$ respectively. Those continuous latent vectors are sampled from a random uniform distribution. They can be converted to $M_{\text{transform}}$  through (8) -- (11). Similarly, training images $X_{\text{real}}$ and transformed images $X_{\text{transform}}$ are encoded to continuous latent vectors through the encoder. They can be further converted to $M_1$ and $M_2$ through equations (8) -- (11). Finally,  the mutual information will be maximized between $M_{\text{transform}}$ and $M_2 * (M_1)^{-1}$ via (15). The updated loss function with affine regularizer is:}

\begin{equation}
\begin{aligned}
&\min_{G,E}\max_{D}V(D,G,E) := \mathbb{E}_{x\sim{P_{\text{data}}(x)}}\log[D(x, E(x))]  \\
&+\mathbb{E}_{{z\sim{P_{z}(z)}},{c\sim{P_{c}(c)}},{c_A\sim{P_{c_A}}(c_A)}}\log[1-D((c,c_A,z), G(c,c_A, z))] \\
&-\lambda I((c,c_A),E(G(c,c_A,z))) - \lambda I(M_{transform}, M_2 * M_1^{-1}) \\\nonumber
\end{aligned}
\end{equation}

\begin{figure*}[htbp]
\centerline{\includegraphics[width = 1\textwidth]{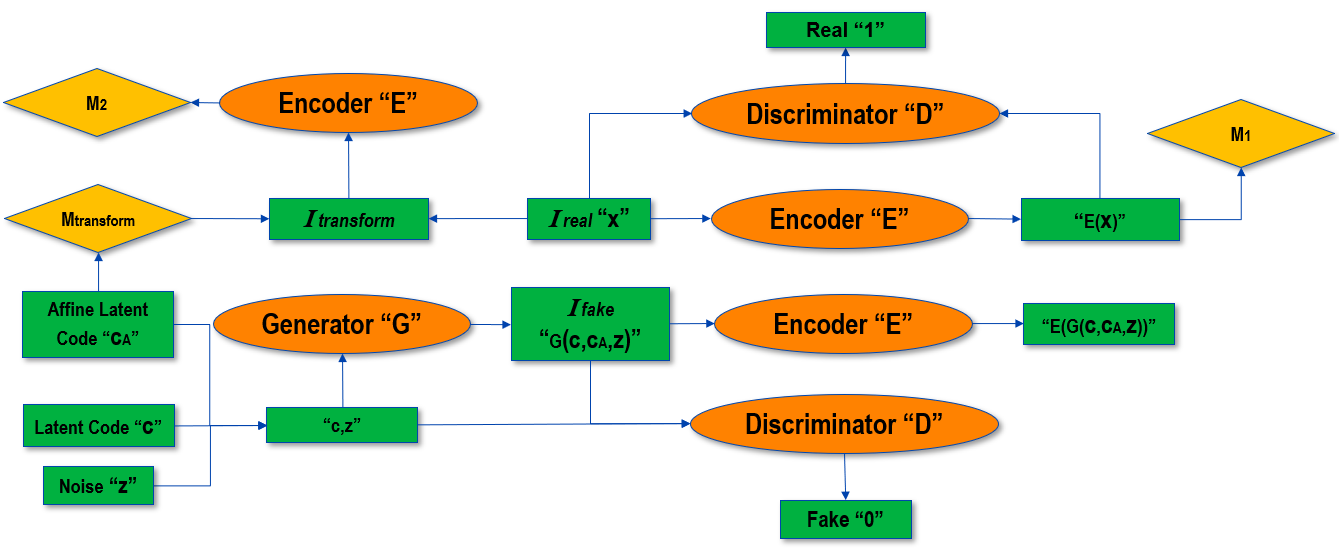}}
\caption{Model architecture.  Diamond boxes are affine regularizers derived in section A. Rectangle boxes are variables. Ellipse boxes are neural networks.}
\end{figure*}

\begin{table}[htbp]
\caption{Layer Information}
\begin{center}
\begin{tabular}{|c|c|c|c|}
\hline
\textbf{}&\multicolumn{3}{|c|}{\textbf{Block}} \\
\cline{2-4} 
\textbf{Layer} & \textbf{\textit{Encoder}}& \textbf{\textit{Generator}}& \textbf{\textit{Discriminator}} \\

\hline
 Input & 28x28 & 72 & 2x28, 72   \\
\hline
NN &  \makecell {3x3x16 conv \\LReLU,dropout} & \makecell{ 128x7x7 fc \\ReLU} & \makecell{3x3x16 conv \\LReLU, dropout, BN} \\
\hline
NN&	\makecell{3x3x32 conv \\LReLU,dropout} &	\makecell{3x3x16 deconv\\ ReLU, BN}&	\makecell{3x3x32 conv \\LReLU, dropout, BN} \\
\hline
NN	&  \makecell{3x3x64 conv \\LReLU,dropout} &	\makecell{3x3x32 deconv\\ ReLU, BN} &	\makecell{3x3x64 conv\\ LReLU,dropout, BN} \\
\hline
NN &	\makecell{3x3x128 conv \\LReLU, dropout\\ BN}	& \makecell{3x3x64 deconv\\ ReLU, BN} &	\makecell{3x3x128 conv \\LReLU,dropout, BN} \\
\hline
NN	&  \makecell{1024 fc\\ LReLU}	& \makecell{ 3x3x128 deconv \\ReLU, BN} &	\makecell{1024 fc \\ LReLU} \\
\hline
NN	& -	& \makecell{3x3x1 conv\\ tanh}	& \makecell{1024 fc\\ LReLU} \\
\hline
NN	& -	& - &	\makecell{1 fc\\ sigmoid} \\
\hline
Output &	\makecell{categorical: 10 fc\\ softmax} &	image: 28x28 &	real/fake: 1 \\
\hline
Output &	continuous: 3 fc & - & -   \\
\hline
Output &	\makecell{noise: 59 fc\\ tanh} & - & - \\
\hline
\multicolumn{4}{l}{ \makecell{NN for neural network, fc for fully connected, conv for convolution, \\deconv for deconvolution, BN for batch normalization}}

\end{tabular}
\label{tab1}
\end{center}

\end{table}

\section{Experimental Results}
{As a proof of concept experiment, we test our algorithm on the MNIST dataset \cite{lecun-mnisthandwrittendigit-2010}. In Section A, we consider experiments with rotated images. In Section B, we explore adversarial attacks. In Section C, we elaborate on the interpretability of the proposed algorithm.}

\subsection{Classification Accuracy on Rotated Images}

{To test the robustness of model against the rotated images, we purposely rotate the images from -30 to +30 degrees as input images. Six models are tested, which are model trained with original dataset, model trained with rotation augmented dataset, model trained with FGSM adversarial sample augmented dataset, model trained with PGD adversarial sample augmented dataset, Bi-directional Info GAN trained with original dataset and the proposed ADIS-GAN trained with original dataset. Adding FGSM and PGD adversarial sample augmented model is to illustrate the robustness of adversarial training against rotated images. Adding Bi-directional Info GAN is to illustrate the robustness of generative model without affine inductive bias against rotated images. \par}

\begin{figure}[htbp]
\centerline{\includegraphics[width = 0.50\textwidth]{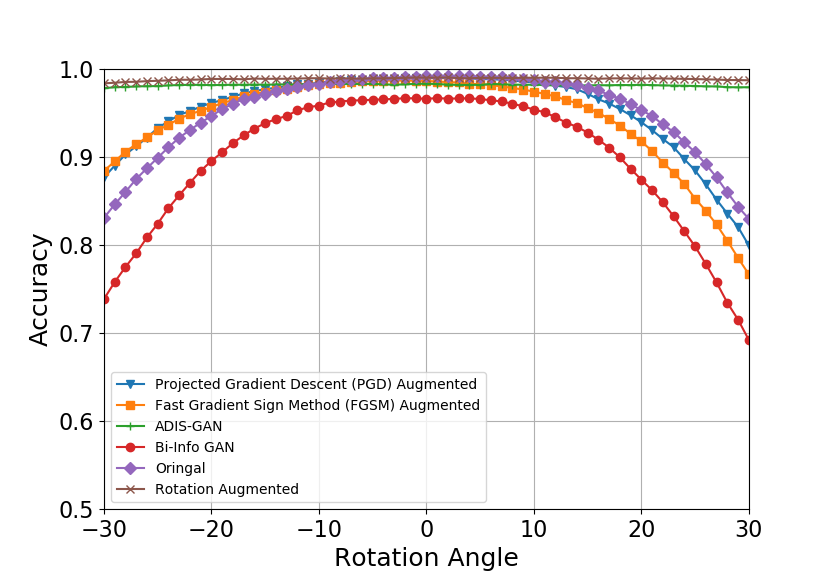}}
\caption{We purposely rotate the images from MNIST test dataset from -30 degrees to +30 degrees as the input images. Affine Disentangled GAN (ADIS-GAN) has achieved over 98 percent accuracy through all rotated degrees. It has less than 1 percent accuracy difference with the model trained on rotational augmented dataset.}
\end{figure}

{As we can observe from Fig.~5, the model trained with clean dataset and adversarial augmented dataset suffer from rotation transformations. PGD Aug and FGSM Aug show that model trained with adversarial sample augmented dataset is not robust against rotation transformations. Bi-directional Info GAN shows that the generative model without affine inductive bias is not robust against rotation transformations. ADIS-GAN has achieved over 98 percent accuracy through all rotation degrees. It has less than 1 percent accuracy difference with the model trained with rotation augmented dataset. This demonstrates the effectiveness of ADIS-GAN against the rotation transformations.}

\subsection{Classification Accuracy on Adversarial Images}

{To test the robustness of model against adversarial attacks, we create two kinds of adversarial samples with FoolBox \cite{DBLP:journals/corr/RauberBB17}.}

\begin{table}[htbp]
\caption{Classification Accuracy against Adversarial Attack}
\begin{center}
\begin{tabular}{|c|c|c|c|}
\hline
\textbf{}&\multicolumn{3}{|c|}{\textbf{Adversarial Attack}} \\
\cline{2-4} 
\textbf{Model} & \textbf{\textit{No Attack}}& \textbf{\textit{\makecell{FGSM \\ $\epsilon$ = 0.3}}}& \textbf{\textit{\makecell{PGD \\ $\epsilon$ = 0.3}}}  \\

\hline
Original & 99.13$\%$ & 25.88 $\%$ &  0.10 $\%$ \\
\hline
\makecell{Rotation Aug} &  98.98 $\%$ & 11.37 $$\% & 0 \\
\hline
\makecell{FGSM Aug} &	98.60$\%$ & 86.57 $\%$  & 61.01 $\%$\\
\hline
\makecell{PGD Aug} & 98.93 $\%$ &91.55$\%$  & 85.88$\%$  \\
\hline
\makecell{ADIS-GAN} &	98.22$\%$	 & 93.10 $\%$ &  96.53 $\%$  \\
\hline
\multicolumn{4}{l}{ \makecell{For PGD attack, the binary search is set to False for speed purpose.\\The defence method for ADIS-GAN is similar to \cite{DBLP:journals/corr/abs-1805-06605}, \\which acts as a filter before the given classifier.}}

\end{tabular}
\label{tab1}
\end{center}

\end{table}

{From Table \RomanNumeralCaps{2} we can observe that the model trained with clean dataset and rotation augmented dataset are vulnerable to adversarial attacks, which shows that affine transformation augmented data training is orthogonal to adversarial attack. PGD is a relatively stronger attack compared to FGSM, and has a higher attack success rate, which shows that adversarial sample augmented training has its limitation: a stronger attack can defeat a model trained with a weaker attack. ADIS-GAN has consistently good performance with those 2 attacks, which shows it may capture a smoother data distribution that involves larger adversarial manifolds.}

\subsection{Interpretability}
{Affine Disentangled GAN (ADIS-GAN) can express the data distribution in a more interpretable way, which mitigates the black box problem of deep learning to some degree. In this section, mapping between rotation angle and latent vectors is shown to explain how the algorithm understands rotational knowledge. Generated images are shown to demonstrate the relationship between latent space and data space.}

\begin{figure}[htbp]
\centerline{\includegraphics[width = 0.55\textwidth]{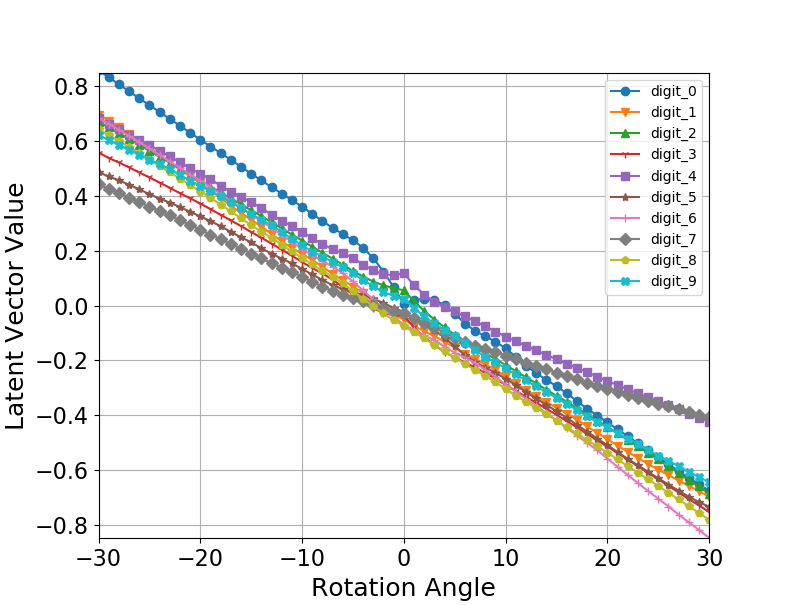}}
\caption{The mapping between rotation angle and latent vector value. The latent vector value has a linear relationship with the rotation angle. The tiny disturbance around 0 degrees is due to the different writing style of hand-written digits.}
\end{figure}

{As we can observe from Fig.~6, the latent vector values have a linear relationship with rotation angle. This explains why ADIS-GAN is robust against image rotation since it can interpret the rotation angles of the given image, which provides information that helps to understand potential misclassifications.  }

\begin{figure}[htbp]
\centerline{\includegraphics[width = 0.52\textwidth]{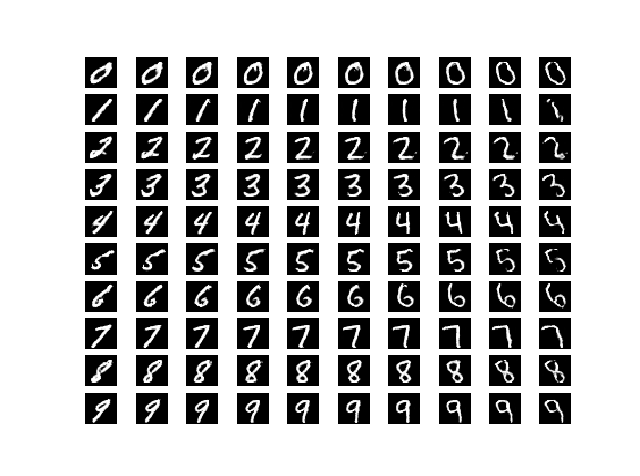}}
\caption{Images change with rotation latent vectors.}
\end{figure}

\begin{figure}[htbp]
\centerline{\includegraphics[width = 0.52\textwidth]{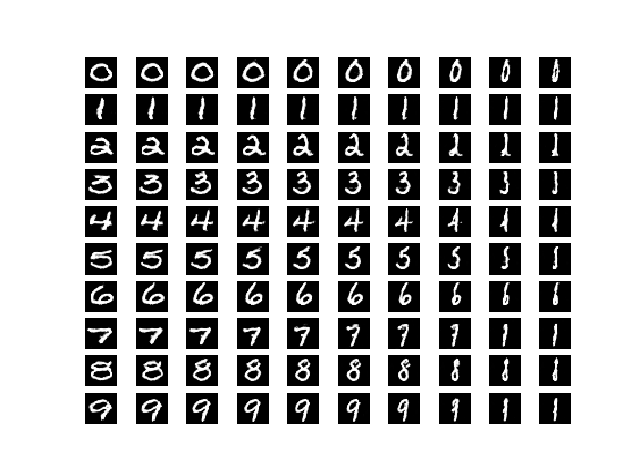}}
\caption{Images change with horizontal zoom latent vectors.}
\end{figure}

{Fig.~7 and Fig.~8 show how the generated images change with latent vectors. These figures illustrate how the algorithm represents information.}

\section{Conclusion and Future Work}
{Deep CNN based vision systems play a major role in the autonomous vehicle (AV) perception system. However, deep CNNs are not robust against affine transformations and adversarial attacks. The former could happen during normal driving scenarios when the vehicle is hitting water puddles or on rural roads, while the latter could happen when a malicious attack is implemented. It is necessary to overcome these challenges before integrating deep CNN based vision system to safety-related applications such autonomous vehicles (AV). \par }
{In this paper, we present the Affine Disentangled GAN (ADIS-GAN), which is robust to both rotation transformations and adversarial attacks. We also introduce the development of affine regularizer. We show that affine transformation augmented and adversarial augmented training is orthogonal, which means they can only defend typical attack they have been trained with. \par}
{Affine regularizer captures the symmetry transformations between latent space and image space during affine transformation. We believe there are many such kinds of symmetries in the physical world. By successfully mapping those symmetries, we can make the deep learning algorithm more robust and interpretable.}

\bibliographystyle{plain}
\bibliography{References}

\begin{appendices}

\section{Two Affine Transformation Orders}
{In principle, there are 6 sequences of affine transformations (R for rotation, K for skew, Z for zoom): \par }
\begin{itemize}
\item{RKZ - 1},
\item{RZK - 1},
\item{KRZ - 2},
\item{KZR - 2},
\item{ZKR - 2},
\item{ZRK - 1}.
\end{itemize}
\hspace{\parindent}{The zoom operation can be inserted arbitrarily in the sequence, since it is a commutative operation. On the other hand, rotation and skew are non-commutative, therefore, their order in the sequence is essential. We can categorize the sequences according to whether the rotation operator is applied before skew or vice versa, leading to two different categories. \par}
{The affine transformation RKZ, which is an example of the first category, can be written as: \par}
\begin{gather}
\begin{bmatrix}A_{11} & A_{12} \\ A_{21} & A_{22}\end{bmatrix}
=
\begin{bmatrix}\cos{\theta} & -\sin{\theta} \\ \sin{\theta} & \cos{\theta} \end{bmatrix}
\begin{bmatrix}1 & m \\ n & 1 \end{bmatrix}
\begin{bmatrix} p & 0 \\ 0 & q \end{bmatrix}.
\end{gather}

{The affine transformation KRZ, which is an example of the second category, can be written as: \par}
\begin{gather}
\begin{bmatrix}A_{11} & A_{12} \\ A_{21} & A_{22}\end{bmatrix}
=
\begin{bmatrix}1 & m \\ n & 1 \end{bmatrix}
\begin{bmatrix}\cos{\theta} & -\sin{\theta} \\ \sin{\theta} & \cos{\theta} \end{bmatrix}
\begin{bmatrix} p & 0 \\ 0 & q \end{bmatrix}.
\end{gather}

{The matrix elements are computed as follows (left: category 1, right: category 2): \par}
\begin{align}
& A_{11} = p(\cos{\theta} - n\sin{\theta}) \longleftrightarrow A_{11} = p(\cos{\theta} + m\sin{\theta}),\\ 
& A_{12} = q(m\cos{\theta} - \sin{\theta}) \longleftrightarrow A_{12} = q(m\cos{\theta} - \sin{\theta}),\\
& A_{21} = p(\sin{\theta} + n\cos{\theta}) \longleftrightarrow A_{21} = p(\sin{\theta} + n\cos{\theta}),\\
& A_{22} = q(m\sin{\theta} + \cos{\theta})  \longleftrightarrow A_{22} = q(-n\sin{\theta} + \cos{\theta}).
\end{align}

{We can observe that A\textsubscript{12} and A\textsubscript{21} are the same for both categories, while A\textsubscript{11} and A\textsubscript{22} are different.}

\end{appendices}

\end{document}